\documentclass[10pt, conference]{IEEEtran}
\IEEEoverridecommandlockouts
\usepackage{cite}
\usepackage{framed,multirow}
\usepackage{graphicx,epsfig}
\usepackage{amssymb,amsmath,bm,amsfonts}
\usepackage{algorithmic}
\usepackage{textcomp}
\usepackage{amsthm}
\usepackage{lineno}
\begin{document}
%
\title{A GRU-based Encoder-Decoder Approach with Attention for Online Handwritten Mathematical Expression Recognition}

\author{\IEEEauthorblockN{Jianshu Zhang\IEEEauthorrefmark{1},
Jun Du\IEEEauthorrefmark{1} and
Lirong Dai\IEEEauthorrefmark{1}}
\IEEEauthorblockA{\IEEEauthorrefmark{1}National Engineering Laboratory for Speech and Language Information Processing\\
University of Science and Technology of China,
Hefei, Anhui, P. R. China\\ Email: xysszjs@mail.ustc.edu.cn, jundu@ustc.edu.cn, lrdai@ustc.edu.cn}
}

\maketitle

\begin{abstract}
In this study, we present a novel end-to-end approach based on the encoder-decoder framework with the attention mechanism for online handwritten mathematical expression recognition (OHMER). First, the input two-dimensional ink trajectory information of handwritten expression is encoded via the gated recurrent unit based recurrent neural network (GRU-RNN). Then the decoder is also implemented by the GRU-RNN with a coverage-based attention model. The proposed approach can simultaneously accomplish the symbol recognition and structural analysis to output a character sequence in LaTeX format. Validated on the CROHME 2014 competition task, our approach significantly outperforms the state-of-the-art with an expression recognition accuracy of 52.43\% by only using the official training dataset. Furthermore, the alignments between the input trajectories of handwritten expressions and the output LaTeX sequences are visualized by the attention mechanism to show the effectiveness of the proposed method.
\end{abstract}


%

\section{Introduction}
Mathematical expressions are indispensable for describing problems and theories in math, physics and many other fields. With the rapid development of pen-based interfaces and tactile devices, people are allowed to write mathematical expressions on mobile devices using handwriting. However, the automatic recognition of these handwritten mathematical expressions is quite different from the traditional character recognition problems with more challenges \cite{Anderson1967,Belaid1984,Miller1998}, e.g., the complicated geometric structures, enormous ambiguities in handwritten input and the strong dependency on contextual information. This study focuses on the online handwritten mathematical expression recognition (OHMER), which attracts broad attention such as the Competition on Recognition of Online Handwritten Mathematical Expressions (CROHME) \cite{ICFHR2014}.

OHMER consists of two major problems \cite{kamfaiChan2000b,Zanibbi2012}, namely symbol recognition and structural analysis, which can be solved sequentially or globally. In the sequential solutions \cite{Zanibbi2002,Alvaro2014}, the errors of symbol recognition and segmentation are subsequently inherited by the structural analysis. Consequently, the global solutions \cite{AwalHarold2014,Alvaro2016} can well address this problem, which are computationally expensive as the probabilities for segmentation composed of strokes are exponentially expanded. Many approaches for structural analysis of mathematical expressions have been investigated, including expression trees \cite{Ha1995}, two-dimensional hidden Markov model (HMM) \cite{Kosmala1998} and others \cite{LeeLee1993,Winkler1995,Nina2015}. Among these, the grammar-based methods \cite{Chou1989,Lavirotte1998} are widely used in OHMER systems \cite{Alvaro2014,AwalHarold2014,Alvaro2016}. These grammars are constructed using extensive prior knowledge with the corresponding parsing algorithms. Overall, both conventional sequential and global approaches have common limitations: 1) the challenging symbol segmentation should be explicitly designed; 2) structural analysis requires the priori knowledge or rules; 3) the computational complexity of parsing algorithms increases exponentially with the size of the predefined grammar.

To address these problems, in this paper, we propose a novel end-to-end approach using the attention based encoder-decoder model with recurrent neural networks (RNNs) \cite{Graves2012} for OHMER. First, the input two-dimensional ink trajectory information of handwritten expression is encoded to the high-level representations via the stack of bi-directional gated recurrent unit based recurrent neural network (GRU-RNN). Then the decoder is implemented by a unidirectional GRU-RNN with a coverage-based attention model \cite{Cho2014,Chorowski2014,Sutskever2014}. The attention mechanism built into the decoder scans the entire input sequence and chooses the most relevant region to describe a segmented symbol or implicit spatial operator. Inherently unlike traditional approaches, our model optimizes symbol segmentation automatically through its attention mechanism, and structural analysis does not rely on a predefined grammar. Moreover, the encoder and the decoder are jointly trained. The proposed encoder-decoder architecture \cite{Bahdanau2015} can make the symbol segmentation, symbol recognition, and structural analysis unified in one data-driven framework to output a character sequence in LaTeX format \cite{latex1985}. Validated on the CROHME 2014 competition task, our approach significantly outperforms the state-of-the-art with an expression recognition accuracy of 52.43\% by only using the official training dataset. Furthermore, the alignments between the input trajectories of handwritten expressions and the output LaTeX sequences are visualized by the attention mechanism to show the effectiveness of the proposed method.

Our proposed approach is related to our previous work \cite{zhang2017watch} and a recent work \cite{Yuntian2016} with the new contributions as: 1) the encoder in this work can fully utilize the online trajectory information via the GRU-RNN while the encoder in \cite{zhang2017watch} using convolutional neural network (CNN) can only work for the offline image as input; 2) different from \cite{Yuntian2016}, the newly added coverage-based attention model is crucial to the recognition performance and its visualization can well explain the effectiveness of the proposed method.

The remainder of the paper is organized as follows. In Section~\ref{proposed}, the details of the proposed approach are introduced. In Section~\ref{exps}, the experimental results and analysis are reported. Finally the conclusion is given in Section~\ref{conclude}.


\section{The Proposed Approach}
\label{proposed}
\begin{figure}
\centering
\includegraphics[width=3.5in]{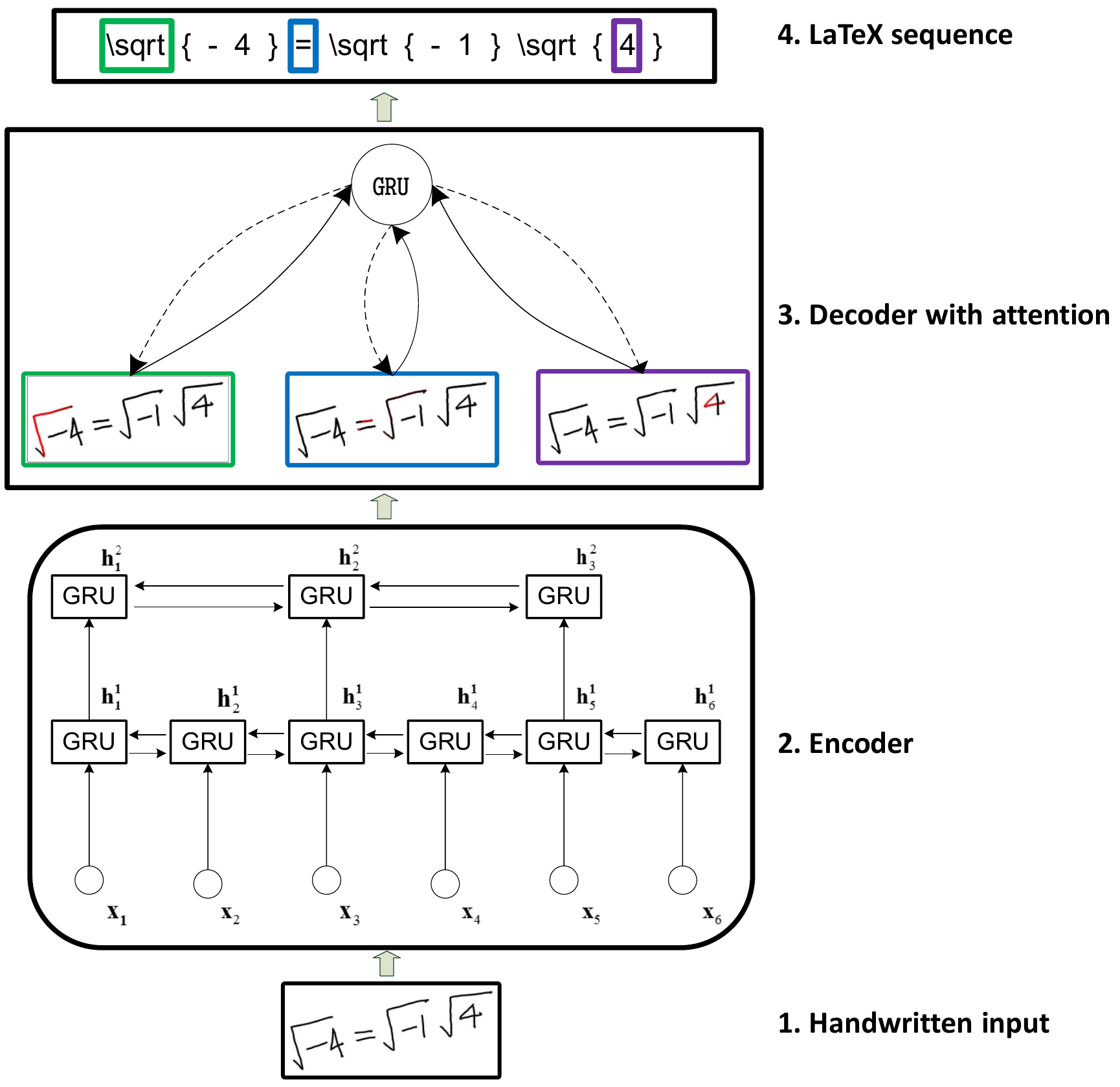}
\caption{The overall architecture of attention based encoder-decoder model.}
\label{fig:overall-architecture}
\end{figure}

In this section, we elaborate the proposed end-to-end framework, namely generating an underlying LaTeX sequence from a sequence of online handwritten trajectory points, as illustrated in Fig. \ref{fig:overall-architecture}. First, the preprocessing is applied to the original trajectory points to extract the input feature vector. Then, the encoder and decoder are well designed using the GRU-RNNs \cite{Chung2014}. The encoder is a stack of bidirectional GRUs while the decoder combines a unidirectional GRU and an attention mechanism into the recurrent sequence generator. The attention mechanism can potentially well learn the alignment between the input trajectory and the output LaTeX sequence. For example in Fig. \ref{fig:overall-architecture}, the green, blue, and purple rectangles denote three symbols with the red color representing the attention probabilities of each handwritten symbol.

\subsection{Preprocessing}
Suppose the input handwritten mathematical expression consists of a sequence of trajectory points with a variable-length $N$:
\begin{equation}\label{eq1}
\left \{\left[ {{x_1},{y_1},{s_1}} \right],\;\left[ {{x_2},{y_2},{s_2}} \right],\; \ldots \;,\;\left[ {{x_N},{y_N},{s_N}} \right] \right\}
\end{equation}
where ${x_i}$ and ${y_i}$ are the xy-coordinates of the pen movements and ${s_i}$ indicates which stroke the $i^{\textrm{th}}$ point belongs to.

To address the issue of non-uniform sampling by different writing speed and the size variations of the coordinates on different potable devices, the interpolation and normalization to the original trajectory points are first conducted according to \cite{xuyao2016}. Then we extract an 8-dimensional feature vector for each point:
\begin{equation}\label{eq2}
\left[ {{x_i},{y_i},\Delta {x_i},\Delta {y_i},{\Delta ^2}{x_i},{\Delta ^2}{y_i},\delta ({s_i} = {s_{i + 1}}),\delta ({s_i} \ne {s_{i + 1}})} \right]
\end{equation}
where $\Delta {x_i}={x_{i+1}}-{x_i}$, $\Delta {y_i}={y_{i+1}}-{y_i}$, ${\Delta ^2} {x_i}={x_{i+2}}-{x_i}$, ${\Delta ^2} {y_i}={y_{i+2}}-{y_i}$ and $\delta ( \cdot ) = 1$ when the condition is true or zero otherwise. The last two terms are flags which indicate the status of the pen, i.e., $\left[ {1,\;0} \right]$ and $\left[ {0,\;1} \right]$ are pen-down and pen-up, respectively. A handwritten mathematical expression is actually composed of several strokes. So fully utilizing the stroke segmentation information plays an important role in constructing an effective recognizer. For convenience, in the following sections, we use $\mathbf{X}=\left( {{\mathbf{x}_1},\;{\mathbf{x}_2},\; \ldots ,\;{\mathbf{x}_N}} \right)$ to denote the input sequence of the encoder, where ${\mathbf{x}_i} \in {\mathbb{R}^d}$ ($d=8$).

\subsection{Encoder}

Given the input sequence $\left( {{\mathbf{x}_1},\;{\mathbf{x}_2},\; \ldots ,\;{\mathbf{x}_N}} \right)$, a simple RNN can be adopted as an encoder to compute the corresponding sequence of hidden state $\left( {{\mathbf{h}_1},\;{\mathbf{h}_2},\; \ldots ,\;{\mathbf{h}_N}} \right)$:
\begin{equation}\label{eq3}
{{\bf{h}}_t} = \tanh \left( {{{\bf{W}}_{xh}}{{\bf{x}}_t} + {{\bf{W}}_{hh}}{{\bf{h}}_{t - 1}}} \right)
\end{equation}
where ${{\bf{W}}_{xh}}$ is the connection weight matrix of the network between input layer and hidden layer, and ${{\bf{W}}_{hh}}$ is the weight matrix of recurrent connections  in a hidden layer. In principle, the recurrent connections can make RNN map from the entire history of previous inputs to each output. However, in practice, a simple RNN is difficult to train properly due to the problems of the vanishing gradient and the exploding gradient as described in \cite{Bengio1994,zhang2016rnn}.

Therefore, in this study, we utilize GRU as an improved version of simple RNN which can alleviate the vanishing and exploding gradient problem. The encoder GRU hidden state ${\mathbf{h}_t}$ is computed as follows:
\begin{align}\label{al1}
 & {{\mathbf{z}}_t} = \sigma ({{\mathbf{W}}_{xz}}{{\mathbf{x}}_{t}} + {{\mathbf{U}}_{hz}}{{\mathbf{h}}_{t - 1}}) \\
 & {{\mathbf{r}}_t} = \sigma ({{\mathbf{W}}_{xr}}{{\mathbf{x}}_{t}} + {{\mathbf{U}}_{hr}}{{\mathbf{h}}_{t - 1}}) \\
 & {{\bf{\tilde h}}_t} = \tanh ({{\bf{W}}_{xh}}{{\bf{x}}_{t}} + {{\bf{U}}_{rh}}({{\bf{r}}_t} \otimes {{\bf{h}}_{t - 1}})) \\
 & {{\bf{h}}_t} = (1 - {{\bf{z}}_t}) \otimes {{\bf{h}}_{t - 1}} + {{\bf{z}}_t} \otimes {{\bf{\tilde h}}_t}
\end{align}
where $\sigma$ is the sigmoid function and $\otimes$ is an element-wise multiplication operator. ${{\mathbf{z}}_t}$, ${{\mathbf{r}}_t}$ and ${{\bf{\tilde h}}_t}$ are the update gate, reset gate and candidate activation, respectively. ${\mathbf{W}}_{xz}$, ${\mathbf{W}}_{xr}$, ${\mathbf{W}}_{xh}$, ${\mathbf{U}}_{hz}$, ${\mathbf{U}}_{hr}$ and ${\mathbf{U}}_{rh}$ are related weight matrices.

Nevertheless, unidirectional GRU cannot utilize the future context. To address this issue, we pass the input vectors through two GRU layers running in opposite directions and concatenate their hidden state vectors. This bidirectional GRU can use both past and future information. To obtain a better representation for the decoder to attend, we stack multiple layers of GRU on top of each other as the encoder. However, as the depth of encoder increases, the high-level representation might contain much redundant information. So we add pooling over time in high-level GRU layers as illustrated by Fig. \ref{fig:overall-architecture}. The pooling operation not only helps accelerate the training process, but also improves the recognition performance as the decoder is easier to attend with a fewer number of outputs of encoder.

\subsection{Decoder equipped with attention mechanism}
As shown in Fig. \ref{fig:overall-architecture}, the decoder generates a corresponding LaTeX sequence of the input handwritten mathematical expression. The output sequence ${\bf{Y}}$ is encoded as a sequence of one-shot vectors.
\begin{equation}\label{eq4}
 \bf{Y} = \left\{ { \mathbf{y}_1, \ldots ,\mathbf{y}_C} \right\}\;,\;{{\mathbf{y}}_i} \in {\mathbb{R}^K}
\end{equation}
where $K$ is the number of total symbols/words in the vocabulary and $C$ is the length of a LaTeX sequence. Meanwhile, the bi-directional GRU encoder produces an annotation sequence ${\bf{A}}$ with a length ${L}$. If there is no pooling in the bi-directional GRU encoder, $L = N$. Each of these annotations is a $D$-dimensional vector:
\begin{equation}\label{eq5}
  \bf{A} = \left\{ {{{\mathbf{a}}_1}, \ldots ,{{\mathbf{a}}_L}} \right\}\;,\;{{\mathbf{a}}_i} \in {\mathbb{R}^D}
\end{equation}

Note that, both the length of annotation sequence $L$ and the length of LaTeX sequence $C$ are not fixed. To address the learning problem of variable-length annotation sequences and associate them with variable-length output sequences, we attempt to compute an intermediate fixed-size vector ${{\mathbf{c}}_t}$, which will be described later. Given the context vector ${{\mathbf{c}}_t}$, we utilize unidirectional GRU to produce the LaTeX sequences symbol by symbol. The probability of each predicted symbol is calculated as:
\begin{equation}\label{eq6}
  p({{\mathbf{y}}_t}|{\mathbf{X}},{{\mathbf{y}}_{t - 1}}) = g \left ({{\mathbf{W}}_o}({\mathbf{E}}{{\mathbf{y}}_{t - 1}} + {{\mathbf{W}}_s}{{\mathbf{s}}_t} + {{\mathbf{W}}_c}{{\mathbf{c}}_t})\right )
\end{equation}
where $g$ denotes a softmax activation function over all the symbols in the vocabulary. ${{\mathbf{s}}_t}$ is the current hidden state of the GRU decoder and ${{\mathbf{y}}_{t-1}}$ represents the previous target symbol. ${{\mathbf{W}}_o} \in {\mathbb{R}^{K \times m}}$, ${{\mathbf{W}}_s} \in {\mathbb{R}^{m \times n}}$, ${{\mathbf{W}}_c} \in {\mathbb{R}^{m \times D}}$, and ${\mathbf{E}}$ denotes the embedding matrix. $m$ and $n$ are the dimensions of embedding and GRU decoder. The GRU decoder also takes the previous target symbol ${{\mathbf{y}}_{t-1}}$ and the context vector ${{\mathbf{c}}_t}$ as input, and employs a single unidirectional GRU layer to calculate the hidden state ${{\mathbf{s}}_t}$:
\begin{align}\label{al2}
 & {{\mathbf{z}}'_t} = \sigma ({{\mathbf{W}}_{yz}}{\mathbf{E}}{{\mathbf{y}}_{t - 1}} + {{\mathbf{U}}_{sz}}{{\mathbf{s}}_{t - 1}} + {{\mathbf{C}}_{cz}}{{\mathbf{c}}_t}) \\
 & {{\mathbf{r}}'_t} = \sigma ({{\mathbf{W}}_{yr}}{\mathbf{E}}{{\mathbf{y}}_{t - 1}} + {{\mathbf{U}}_{sr}}{{\mathbf{s}}_{t - 1}} + {{\mathbf{C}}_{cr}}{{\mathbf{c}}_t}) \\
 & {{\bf{\tilde s}}_t} = \tanh ({{\bf{W}}_{ys}}{\bf{E}}{{\bf{y}}_{t - 1}} + {{\bf{U}}_{rs}}({{\mathbf{r}}'_t} \otimes {{\bf{s}}_{t - 1}}) + {{\bf{C}}_{cs}}{{\bf{c}}_t}) \\
 & {{\bf{s}}_t} = (1 - {{\bf{z}}'_t}) \otimes {{\bf{s}}_{t - 1}} + {{\bf{z}}'_t} \otimes {{\bf{\tilde s}}_t}
\end{align}
where ${{\mathbf{z}}'_t}$, ${{\mathbf{r}}'_t}$ and ${{\bf{\tilde s}}_t}$ are the update gate, reset gate and candidate activation, respectively. ${\mathbf{W}}_{yz}$, ${\mathbf{W}}_{yr}$, ${\mathbf{W}}_{ys}$, ${\mathbf{U}}_{sz}$, ${\mathbf{U}}_{sr}$, ${\mathbf{U}}_{rs}$, ${\mathbf{C}}_{cz}$, ${\mathbf{C}}_{cr}$ and ${\mathbf{C}}_{cs}$ are related weight matrices.

Intuitively, for each predicted symbol from the decoder, not the entire input sequence is useful. Only a subset of adjacent trajectory points should mainly contribute to the computation of context vector ${{\mathbf{c}}_t}$ at each time step $t$. Therefore, the decoder can adopt an attention mechanism to link to the related part of input sequence and then assign a higher weight to the corresponding annotation vector ${{\mathbf{a}}_i}$. Here, we parameterize the attention model as a multi-layer perceptron (MLP) that is jointly trained with the encoder and the decoder:
\begin{equation}\label{eq7}
  {e_{ti}} = {\bm{\nu }}_{att}^{\rm T}\tanh ({{\mathbf{W}}_{att}}{{\mathbf{s}}_{{t - 1}}} + {{\mathbf{U}}_{att}}{{\mathbf{a}}_{i}})
\end{equation}
\begin{equation}\label{eq8}
  {\alpha _{ti}} = \frac{{\exp ({e_{ti}})}}{{\sum\nolimits_{k = 1}^L {\exp ({e_{tk}})} }}
\end{equation}
Let $n^{'}$ denote the attention dimension. Then ${{\bm{\nu }}_{att}} \in {\mathbb{R}^{{n^{'}}}}$, ${{\mathbf{W}}_{att}} \in {\mathbb{R}^{{n^{'}} \times n}}$ and ${{\mathbf{U}}_a} \in {\mathbb{R}^{{n^{'}} \times D}}$. With the weights ${\alpha _{ti}}$, the context vector ${{\mathbf{c}}_t}$ is calculated as:
\begin{equation}\label{eq9}
  {{\mathbf{c}}_t} = \sum\nolimits_i^L {{\alpha _{ti}}{{\mathbf{a}}_i}}
\end{equation}

The attention probability ${\alpha _{ti}}$ denotes the alignment between the target symbol and a local region of input sequence. It can also be considered as a regularization parameter for the bi-directional GRU encoder because the attention helps diminish the gradient back-propagated from the decoder.

\subsection{Coverage based attention model}
However, there is one problem for the conventional attention mechanism in \eqref{eq7}, namely the lack of coverage \cite{lihang2016}. Coverage represents overall alignment information. An attention model lacking coverage is not aware whether a part of input expression has been translated or not. Misalignment will lead to over-translating or under-translating. Over-translating means that some parts of the input sequence have been translated twice or more, while under-translating implies that some parts have never been translated. To address this problem, we append a coverage vector to the computation of attention in \eqref{eq7}. The coverage vector aims at providing alignment information. Different from \cite{Bahdanau2016}, we compute the coverage vector based on the sum of all past attention probabilities ${{\bm{\beta}}_t}$, which can describe the alignment history:
\begin{align}\label{al3}
  & {{\bm{\beta}}_t} = \sum\nolimits_{l}^{t - 1} {{{\bm{\alpha}}_l}} \\
  & {\mathbf{F}} = {\mathbf{Q}} * {{\bm{\beta }}_t} \\
  & {e_{ti}} = {\bm{\nu }}_{att}^{\rm T}\tanh ({{\mathbf{W}}_{att}}{{\mathbf{s}}_{t - 1}} + {{\mathbf{U}}_{att}}{{\mathbf{a}}_i} + {{\mathbf{U}}_f}{{\mathbf{f}}_i})
\end{align}
where ${{{\bm{\alpha}}_l}}$ is the attention probability vector at time step $l$ and ${{\mathbf{f}}_i}$ denotes the $i^{\textrm{th}}$ coverage vector of ${\mathbf{F}}$. ${{\bm{\beta}}_t}$ is initialized as a zero vector.  The coverage vector is produced through a convolutional layer because we believe the coverage vector of annotation ${{\mathbf{a}}_i}$ should also be associated with its adjacent attention probabilities.

The coverage vector is expected to adjust the future attention. More specifically, trajectory points in the input sequence already significantly contributed to the generation of target symbols should be assigned with lower attention probabilities in the following decoding phases. On the contrary, trajectory points with less contributions should be assigned with higher attention probabilities. Consequently, the decoding process is finished only when the entire input sequence has contributed and the problems of over-translating or under-translating can be alleviated.

\section{Experiments}
\label{exps}

The experiments are conducted on CROHME 2014 competition dataset. The training set consists of 8836 handwritten mathematical expressions (about 86000 symbols) while the test set includes 986 expressions (about 6000 symbols). There are totally 101 maths symbol classes. None of the expressions in the test set is seen in the training set. To be fairly comparable, we also used the CROHME 2013 test set as a validation set in the training stage, just like other participants of CROHME 2014 competition.

The training objective of our model is to maximize the predicted symbol probability as shown in \eqref{eq6} and we use cross-entropy (CE) as the criterion. The encoder consists of 4 layers of bi-directional GRUs. Each layer has 250 forward and 250 backward GRU units. The pooling is applied to the top 2 GRU layers over time. Accordingly, the encoder reduces the input sequence length by the factor of 4. The decoder is a single layer with 256 forward GRU units. The embedding dimension $m$ and GRU decoder dimension $n$ are set to 256. The attention dimension $n'$ and annotation dimension $D$ are set to 500.  We utilize the AdaDelta algorithm \cite{adadelta2012} with gradient clipping for optimization. The AdaDelta hyperparameters are set as $\rho  = 0.95$, $\varepsilon  = {10^{ - 6}}$. The early-stopping of training procedure is determined by word error rate (WER) \cite{WER2002} of validation set. We use the weight noise \cite{weightnoise2011} as the regularization. The training is first finished without weight noise, we then anneal the best model in terms of WER by restarting the training with weight noise.

In the decoding stage, we aim to generate a most likely LaTeX sequence given the input sequence. The beam search algorithm \cite{cho2015BM} is employed to complete the decoding process. At each time step, we maintain a set of 10 partial hypotheses. We also adopt the ensemble method \cite{ensemble} to improve the performance of our neural network model. We first train 5 models on the same training set but with different initialized parameters and then average their prediction probabilities on the generated symbol during the beam search process.

\subsection{Recognition performance}
\begin{table}[h]
\caption{\label{tab:1}{Correct expression recognition rates (in \%) of different systems on CROHME 2014 test set.}}
\centering
\begin{tabular}{c c c c c}
\hline
\textbf{System} & \textbf{Correct(\%)} & \textbf{$\leq$ 1(\%)} & \textbf{$\leq$ 2(\%)} & \textbf{$\leq$ 3(\%)}\\
\hline
\uppercase\expandafter{\romannumeral1} & 37.22 & 44.22 & 47.26 & 50.20 \\
\uppercase\expandafter{\romannumeral2} & 15.01 & 22.31 & 26.57 & 27.69 \\
\uppercase\expandafter{\romannumeral4} & 18.97 & 28.19 & 32.35 & 33.37 \\
\uppercase\expandafter{\romannumeral5} & 18.97 & 26.37 & 30.83 & 32.96 \\
\uppercase\expandafter{\romannumeral6} & 25.66 & 33.16 & 35.90 & 37.32 \\
\uppercase\expandafter{\romannumeral7} & 26.06 & 33.87 & 38.54 & 39.96 \\
\hline
P1 & \textbf{42.49} & \textbf{57.91} & \textbf{60.45} & \textbf{61.56} \\
P2 & \textbf{46.86} & \textbf{61.87} & \textbf{65.82} & \textbf{66.63} \\
P3 & \textbf{52.43} & \textbf{68.05} & \textbf{71.50} & \textbf{72.31} \\
\hline
\end{tabular}
\end{table}

The comparison among the proposed approach (systems P1, P2, P3) and others on CROHME 2014 test set is listed in Table \ref{tab:1}. Systems \uppercase\expandafter{\romannumeral1} to \uppercase\expandafter{\romannumeral7} were submitted systems to CROHME 2014 competition. Note that system \uppercase\expandafter{\romannumeral3} is not given for a fair comparison as it used additional training data not provided officially. The details of these systems can be seen in \cite{ICFHR2014}. System \uppercase\expandafter{\romannumeral1} acquired an expression rate (ExpRate) of 37.22\% and was awarded the first place on CROHME 2014 competition using the official training data. It should be indicated that there is a large performance gap between the ExpRate of the first place and the second place.

System P1 and P2 are two of our proposed systems without/with coverage based attention model, respectively. System P3 is our best ensemble system with 5 models. It is clear that system P1 even without coverage model can achieve an ExpRate of 42.49\%, which significantly outperforms the best submitted system to CROHME 2014 competition with an absolute gain of about 5\%. By using the coverage model, an absolute gain of 4\% could be obtained from system P1 to P2. The best system P3 yields an ExpRate of 52.43\%, which should be the best published result on CROHME 2014 test set, to the best of our knowledge.

A mathematical expression is considered to be correctly recognized only when the generated LaTeX sequence matches ground truth. Additionally, Table \ref{tab:1} also shows the expression recognition accuracies with one, two and three errors per expression, represented by $(\leq 1)$, $(\leq 2)$ and $(\leq 3)$. The performance gap between correct and error $(\leq 1)$ shows that the corresponding systems still have a large room to be improved. Meanwhile, the differences between error $(\leq 2)$ and error $(\leq 3)$ show that it is difficult to improve the accuracy by incorporating a single correction when more errors happen.

\subsection{Attention visualization}
\begin{figure}
\centering
\includegraphics[width=3.0in]{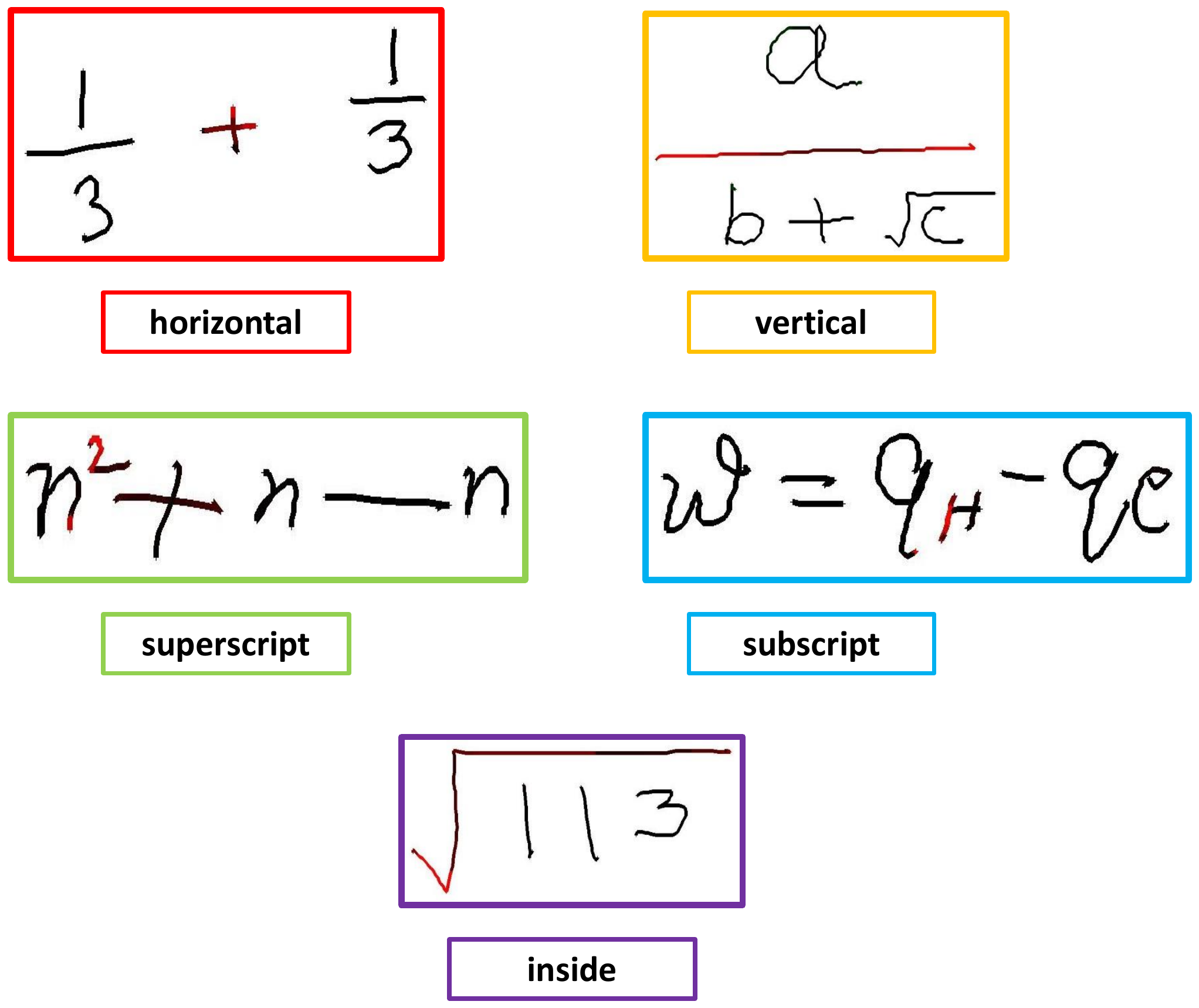}
\caption{The learning of five spatial relationships (horizontal, vertical, subscript, superscript and inside) through attention visualization.}
\label{fig:5spatial}
\end{figure}

\begin{figure}
\centering
\includegraphics[width=3.5in]{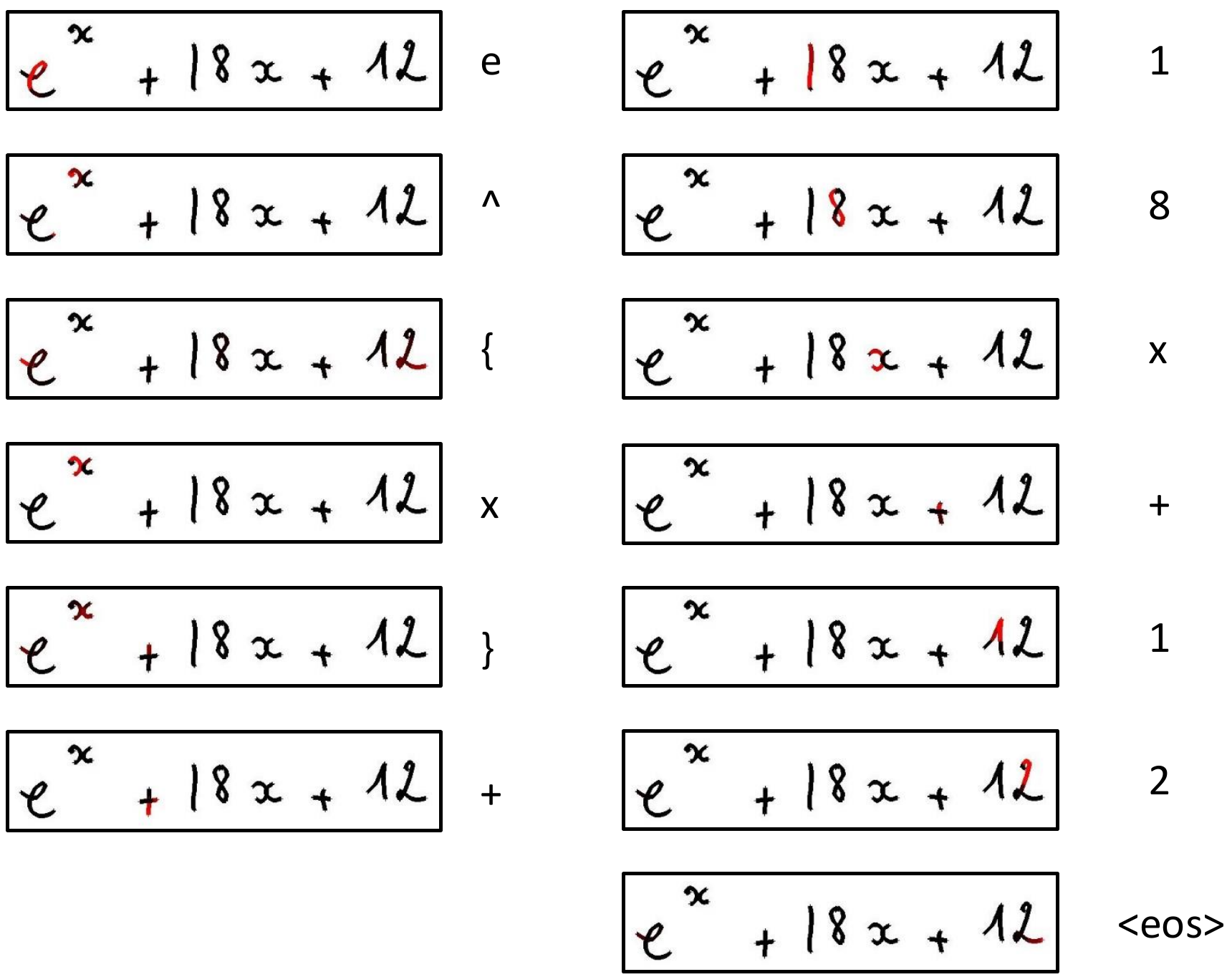}
\caption{Attention visualization for an example of a handwriting mathematical expression with the LaTeX ground truth `` e $\wedge$ \{ x \} + 1 8 x + 1 2 ''.}
\label{fig:attention-example}
\end{figure}
In this section, we show through attention visualization how the proposed model is able to analyse the two-dimensional structure grammar and perform symbol segmentation implicitly. We draw the trajectory of input handwritten mathematical expression in a 2-D image to visualize attention. We use the red color to describe the attention probabilities, namely the higher attention probabilities with the lighter color and the lower attention probabilities with the darker color.

In the two-dimensional grammar of mathematical expressions, there are mainly five kinds of spatial relationship between maths symbols, including horizontal, vertical, subscript, superscript and inside relationships. Correctly recognizing these five spatial relationships is the key to analyse the two-dimensional grammar. As shown in Fig. \ref{fig:5spatial}, the horizontal and vertical relationships are easy to learn by focusing on the middle operator. When dealing with superscripts, the decoder precisely pays attention to the end of base symbols and the start of superscript symbols. It does make sense because trajectory points in the start of superscript symbols are on the upper-right of trajectory points in the end of base symbols, describing the upper-right direction. Similarly, for subscripts, the ending points of base symbols and the starting points of superscript symbols can also describe the bottom-right direction. As for the inside relationships, the decoder attends to the bounding symbols.

More specifically, in Fig. \ref{fig:attention-example}, we take the expression ${e^x} + 18x + 12$ as a correctly recognized example. We show that how our model learns to translate this handwritten mathematical expression from a sequence of trajectory points into a LaTeX sequence `` e $\wedge$ \{ x \} + 1 8 x + 1 2 '' step by step. When encountering basic math symbols like ``e'', ``x'', ``+'', ``1'', ``2'' and ``8'', the attention model well generates the alignment strongly corresponding to the human intuition. When encountering a spatial relationship in ${e^x}$, the attention model correctly distinguishes the upper-right direction and then produces the symbol ``$\wedge$". More interestingly, immediately after detecting the superscript spatial relationship, the decoder successfully generates a pair of braces ``\{\}'', which are used to compose the exponent grammar in LaTeX. Finally, the decoder attends both the end and the start of the entire input sequence and generates an end-of-sentence (eos) mark.

\subsection{Error Analysis}

\begin{figure}
\centering
\includegraphics[width=3.5in]{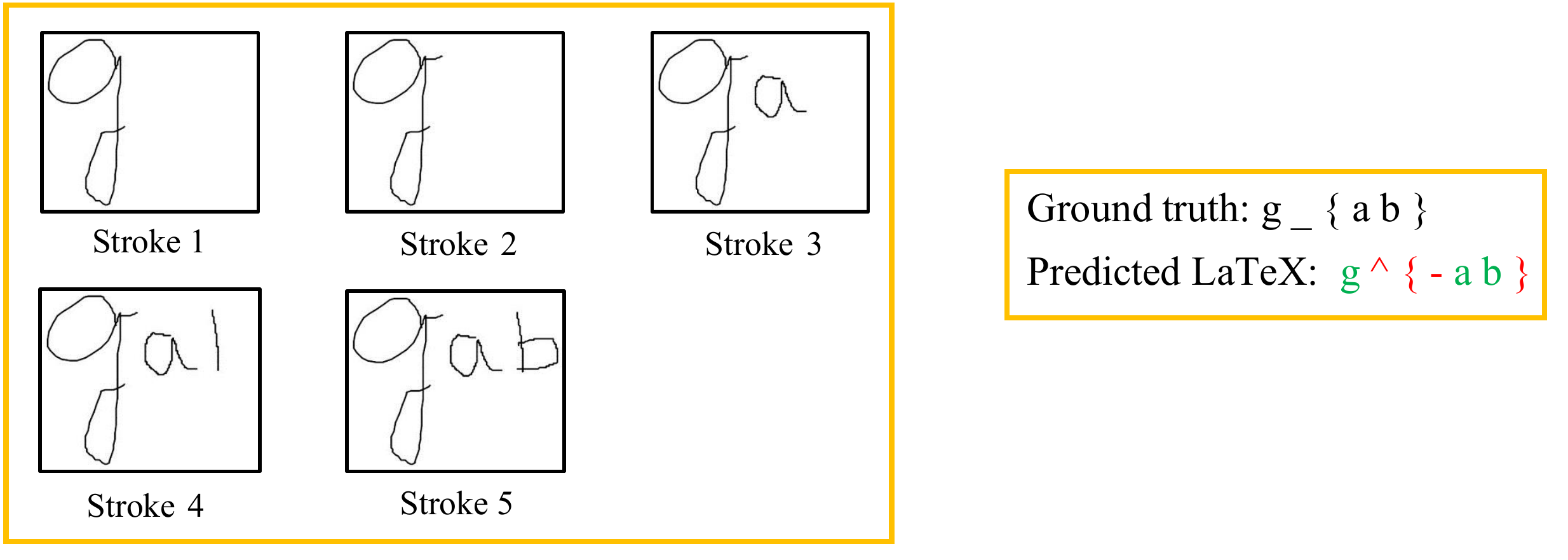}
\caption{Analysis of an incorrectly recognized example of handwritten mathematical expression due to the over-translating problem where ``$\wedge$'' is over-translated.}
\label{fig:over-trans}
\end{figure}

\begin{figure}
\centering
\includegraphics[width=3.5in]{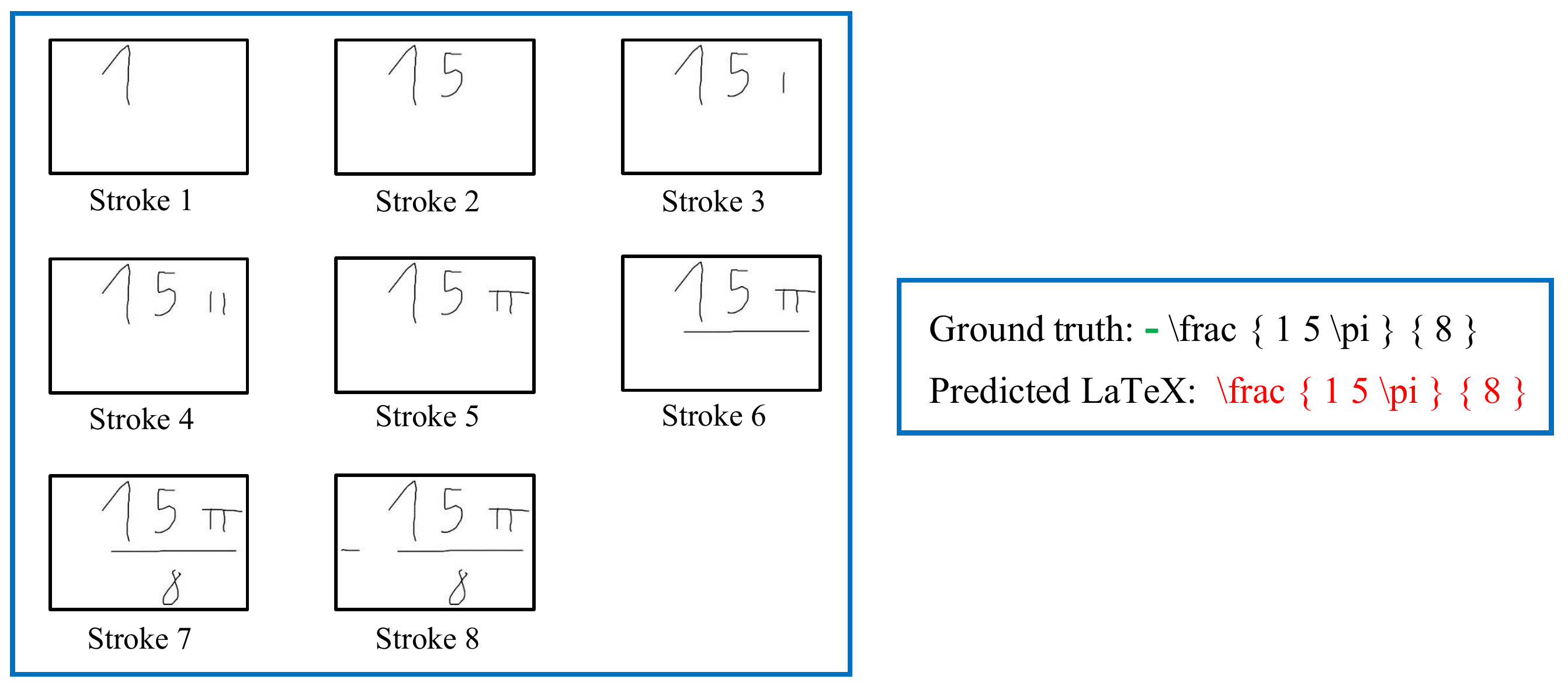}
\caption{Analysis of an incorrectly recognized example of handwritten mathematical expression due to the under-translating problem where the minus sign ``--'' is under-translated.}
\label{fig:under-trans}
\end{figure}

In Fig. \ref{fig:over-trans} and Fig. \ref{fig:under-trans}, we show two typical incorrectly recognized examples of handwritten mathematical expressions, due to over-translating and under-translating, respectively. The stroke 2 in Fig. \ref{fig:over-trans} is an inserted stroke, which is actually the end of symbol ``g'' but split into another stroke by the writer. Accordingly, our model over-translates the input sequence and recognizes the stroke 2 as a minus sign ``--''. And the spatial relationship subscript is mistaken as the superscript. In Fig. \ref{fig:under-trans}, the first symbol of formula LaTeX string, namely the minus sign ``--'', is missing, which corresponds to the last stroke of the handwritten example. In general, we should write the minus sign as the first stroke. Consequently, this inverse stroke leads to the under-translating problem.

\section{Conclusion}
\label{conclude}
In this study we introduce an encoder-decoder with coverage based attention model to recognize online handwritten mathematical expressions. The proposed approach can fully utilize the online trajectory information via GRU-RNN based encoder. And the coverage model is quite effective for attention by using the alignment history information. We achieve promising recognition results on CROHME 2014 competition. We show from experiment results that our model is capable of performing symbol segmentation automatically and learning to grasp a maths grammar without priori knowledge. Also, we demonstrate through attention visualization that the learned alignments by attention model well correspond to human intuition. As for the future work, we aim to improve our approach to reduce the over-translating and under-translating errors.

\section*{Acknowledgment}
The authors want to thank Junfeng Hou for insightful comments and suggestion. This work was supported in part by the National Natural Science Foundation of China under Grants 61671422 and U1613211, in part by the Strategic Priority Research Program of the Chinese Academy of Sciences under Grant XDB02070006, and in part by the National Key Research and Development Program of China under Grant 2016YFB1001300.



%
\bibliographystyle{IEEEtran}
\bibliography{refs}

\end{document}